\documentclass[journal]{IEEEtran}
\usepackage{CJK}
\usepackage{amssymb}
\usepackage{amsmath}
\usepackage{graphicx, subfigure}
\usepackage{url, cite}
\usepackage{verbatim}
\usepackage{booktabs}
\usepackage{multirow}
\usepackage{bigstrut}
\usepackage{booktabs}
\usepackage[ruled]{algorithm2e}
\usepackage{subfigure}
\usepackage{graphicx}

\usepackage{amsmath,cite,graphicx,times,subfigure,url,verbatim}
\usepackage{algorithmic}
\usepackage{epstopdf}
\usepackage{textcomp}
\usepackage{marvosym}
\usepackage{tabularx}

\newtheorem{subsec:coding}{subsec:coding}

\usepackage{color}

\begin{document}

\title{DeepNetQoE: Self-adaptive QoE Optimization Framework of Deep Networks}

\author{
Rui~Wang,~Min~Chen\textsuperscript{\Letter},~Nadra~Guizani,~Yong~Li,~Hamid~Gharavi,~Kai~Hwang
\thanks{R. Wang is with School of Computer Science and Technology, Huazhong University of Science and Technology, China. (ruiwang2018@hust.edu.cn)}
\thanks{M. Chen is with School of Computer Science and Technology and Wuhan National Laboratory for Optoelectronics£¬Huazhong University of Science and Technology. (minchen2012@hust.edu.cn)}
\thanks{N. Guizani is with Department of Electrical and Computer Engineering, University of Idaho, USA. (nguizani@ieee.org)}
\thanks{Y. Li is with the Tsinghua National Laboratory for Information Science and Technology, Department of Electronic Engineering, Tsinghua University, China. (liyong07@tsinghua.edu.cn)}
\thanks{H. Gharavi is with National Institute of Standards and Technology (NIST), Gaithersburg, USA. (hamid.gharavi@nist.gov)}
\thanks{K. Hwang is with The Chinese University of Hong Kong, Shenzhen, China. (hwangkai@cuhk.edu.cn)}
\
\thanks{\textsuperscript{\Letter} Corresponding author.}
}

\markboth{}{}
\markboth{Accepted for publication: IEEE Network}{}
\maketitle

\begin{abstract}
Future advances in deep learning and its impact on the development of artificial intelligence (AI) in all fields depends heavily on data size and computational power. Sacrificing massive computing resources in exchange for better precision rates of the network model is recognized by many researchers. This leads to huge computing consumption and satisfactory results are not always expected when computing resources are limited. Therefore, it is necessary to find a balance between resources and model performance to achieve satisfactory results. This article proposes a self-adaptive quality of experience (QoE) framework, DeepNetQoE, to guide the training of deep networks. A self-adaptive QoE model is set up that relates the model's accuracy with the computing resources required for training which will allow the experience value of the model to improve. To maximize the experience value when computer resources are limited, a resource allocation model and solutions need to be established. In addition, we carry out experiments based on four network models to analyze the experience values with respect to the crowd counting example. Experimental results show that the proposed DeepNetQoE is capable of adaptively obtaining a high experience value according to user needs and therefore guiding users to determine the computational resources allocated to the network models.
\end{abstract}


\IEEEpeerreviewmaketitle

\section{Introduction}
In recent years, the rapid growth of data volume and the significant improvement of computing chips technology have greatly promoted the development of deep networks and the further impact of artificial intelligence (AI). At the same time, deep learning techniques have been widely used in many scenarios. From computer vision to natural language processing, speech recognition to emotion recognition, deep learning has demonstrated its influence even in areas such as autonomous driving and virtual assistants. With the huge non-linear fitting capacity of deep networks being potentially stronger than human beings, it makes it a hotspot in many fields of research~\cite{liunet}.

Nevertheless, a deep neural network is a complicated process that requires extensive training data and strong computing capacity. In a deep neural network, there exists many neurons and the connections among them are huge. Moreover, the foundation of a deep network obtained through iterative training of a dataset and continuously optimizing all parameters until the optimal model (e.g., meeting the needs of the scene) is obtained requires a great deal of investment in computational power. In other words, the stronger the computing power, the faster to obtain optimal values of the deep learning model. Qinrang et al.~\cite{Fpga} propose an implementation method of an FPGA-based convolutional neural network accelerator. The authors proposed approach is aimed at overcoming real-time limitations of the convolutional neural network (CNN) in the embedded field and the sparseness of convolutional calculation of CNN to increase the calculation speed. Vivienne et al.~\cite{VivPro} offers reducing the computational cost of deep neural networks by combining hardware design and deep neural network algorithms. Google~\cite{JouppiTPU} proposes a Tensor Processing Unit (TPU) based architecture, which accelerates the inference phase of the neural network. Many researchers are committed to improving the hardware computing power used for deep network training.
\begin{figure*}
\centering
\includegraphics[width=6.8in]{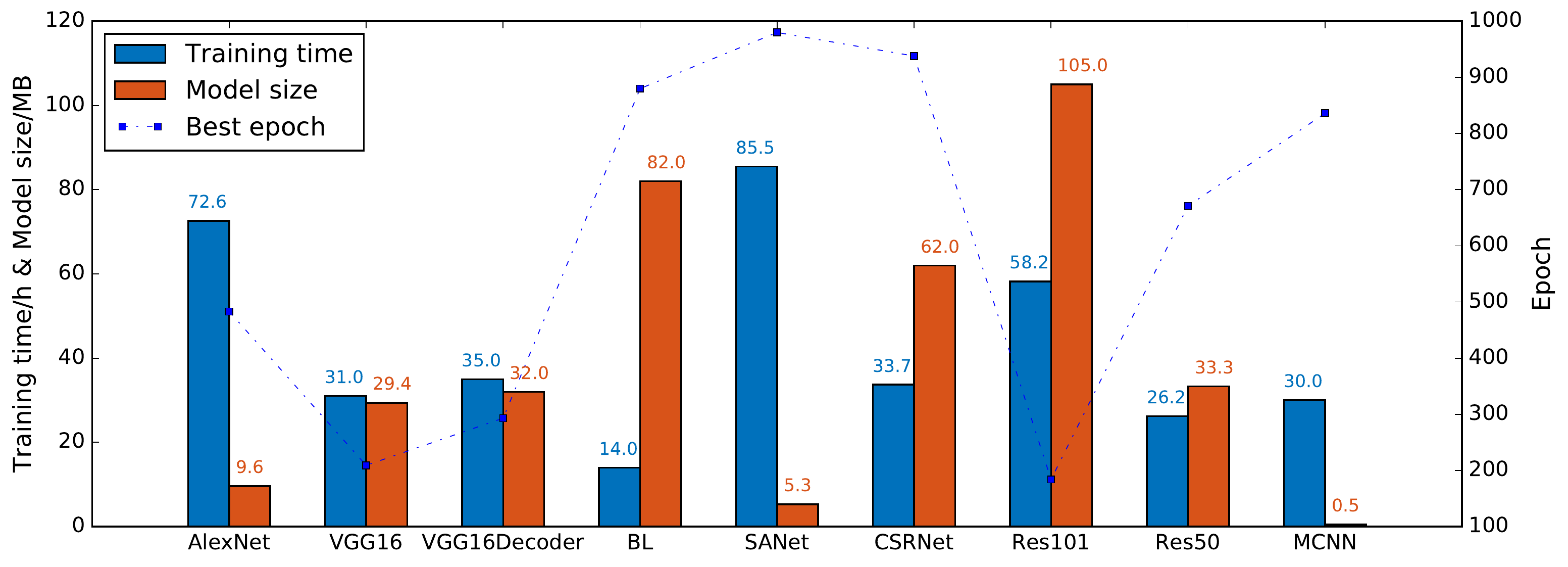}
\caption{The training time and model size of the crowd counting model, and best epoch where the optimal parameters appear.}
\label{fig001}
\end{figure*}

However, the computing resources required during the training process cannot be accurately estimated. This article takes crowd counting as an example, as shown in Figure 1. In the crowd counting task, different networks have been trained at 1000 epochs to obtain training time, model size, and an optimal model iteration number in the same dataset. The degree of investment in computing resources is replaced by training time. As shown in Figure 1, the size of the SANet~\cite{SANet} model size is only 5.3 Mbyte (MB), which requires 85.51 hours (h) of training time. For the Bayesian Loss (BL)~\cite{BL}, training time is only 14 h, but the model size is 82.0 MB. The weak correlation between training time and network parameters adds to the difficulty of predicting the computational power investment during model training. At the same time, to improve the performance of deep networks, researchers often increase the number of training iterations in exchange for higher accuracy, which leads to the consumption of more computing resources. For some models, performance will be further improved as the number of iterations increase, hence making investment of computing power worthwhile. However, this method will fail on some models. As shown in Figure 1 we notice that VGG16, VGG16 Decoder, and Res101~\cite{C3F} follow-up training is meaningless. On the other hand, the performances of BL, SANet, CSRNet~\cite{CSRNet} and MCNN~\cite{MCNN} models continue to improve as iterations increase, and the investment of computing power has achieved an extremely high cost performance. Therefore, in the case of limited computing resources, it is necessary to consider adaptive training methods for performance and resource optimization. In this way, we can accommodate as many calculations and services as possible while improving model training efficiency.

To achieve a balance between the consumption of computing resources and the performance of the model, it is necessary to find an optimal model with a reasonable consumption of resources. For example, in the design of an autonomous mobile robot, Lahijianian et al.~\cite{LahijanianRobot} argue that reducing the consumption of computational resources does not seriously impact the autonomous ability of the robot. This clearly indicates that there is a reasonable tradeoff between resources and performance. Zhang et al.~\cite{ZhangPSAC} develop a robust and effective proactive content caching strategy based on deep learning for improving user experience and reducing network load. Though it could not provide optimal results in consumption, it definitely has practical value for raising the networking service quality. Moreover, there are a few researchers focusing on the allocation of the user's resource request when several different types of resources coexist~\cite{BegaDeepCog}. In fine-grained tasks referring to specific networks to be trained toward GPU resources allocation, related problems and corresponding solutions are in short supply. There are two solutions that can reduce computing resources effectively without impairing the model performance: one is to reduce the running time by modifying parameters in the process of model training and the second, is to provide a resource allocation plan that satisfies the user's quality of experience (QoE) self-adaptively. The latter is based on indexes generated in the model training process, the users' expected model precision, and authorized computing resources~\cite{MarathePer}. The dynamic management of computing resources is particularly important and needs to adapt to the changing service demand over time. However, the following research challenges still exist:
\begin{itemize}
  \item \textbf{Lack of processing flow for deep network model training process optimization}. Due to uncertainty of the deep network training process, there are currently no thorough procedures for optimizing a deep network training process. This produces an inaccuracy in guiding computing resource investments in model training.
  \item \textbf{Lack of fine-grained allocation of computing resources for deep models}. No fine-grained model schemes are currently available in terms of computing resources and computing tasks. Therefore, there are still issues such as unclear indexes and parameters in the actual model allocation plan, which brings more uncertainty to the decision-making plan.
  \item \textbf{Failure to consider users' expectation on performance and resource conditions in the training process}. It is necessary to allocate resources reasonably based on the user's expectation of the performance and computing resources possessed by the user. Nonetheless, the existing research fails to consider the user's needs to achieve the self-adaptivity of QoE.
\end{itemize}

\begin{figure*}
\centering
\includegraphics[width=6.8in]{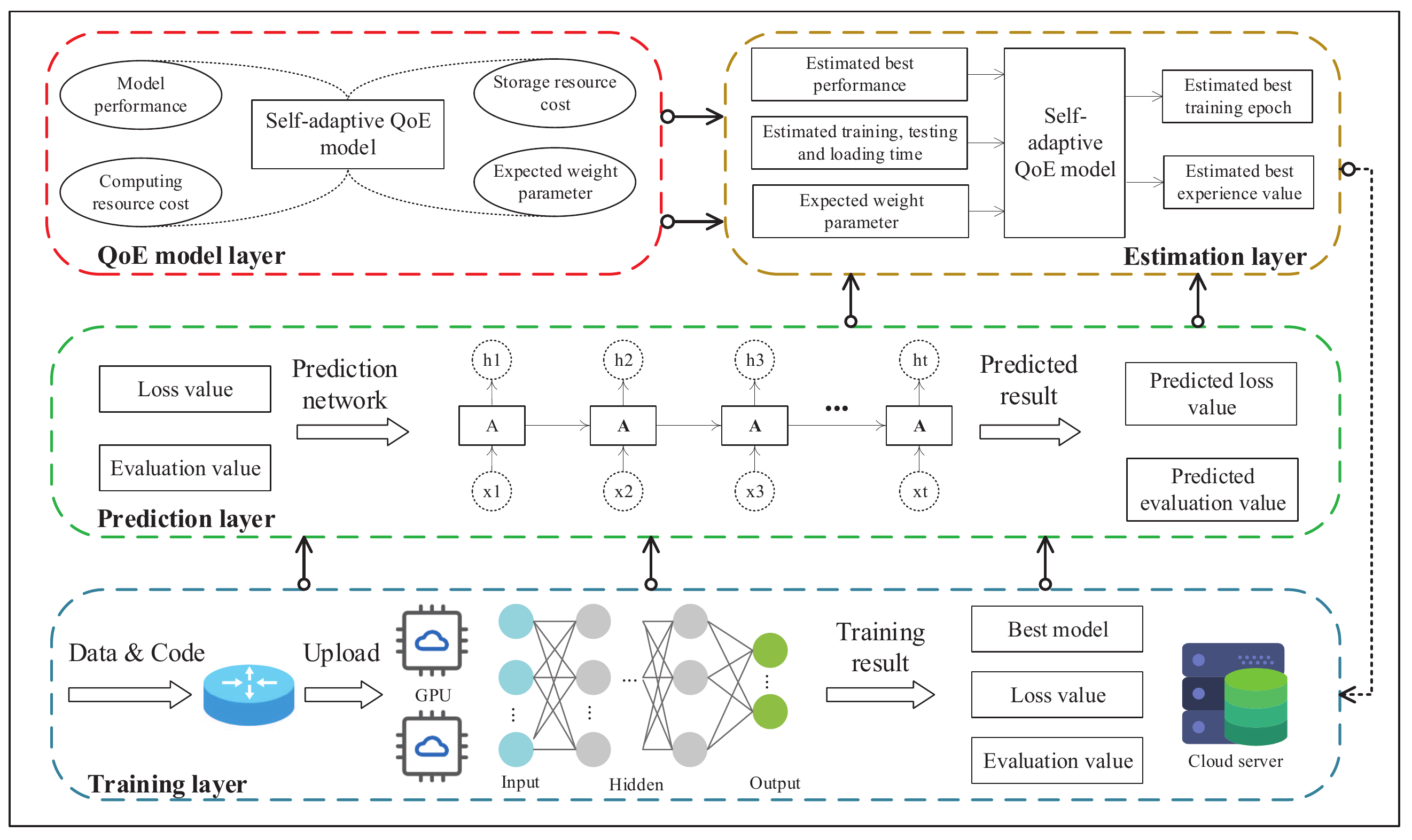}
\caption{The architecture of DeepNetQoE.}
\label{fig002}
\end{figure*}

Based on the shortcomings in current research, this article proposes DeepNetQoE, a self-adaptive QoE optimization framework for deep networks. It combines DeepNetQoE with specific applications to verify its performance in the training of crowd counting models. Driven by the prediction on model performance, the fine-grained computing resources allocation plan of deep model is presented, all while ensuring user needs are satisfied. Therefore, the article has the following contributions:
\begin{itemize}
  \item \textbf{Presents a self-adaptive QoE optimization framework for deep network model training}. In view of the huge consumption of computing resources in the process of deep network training, we propose a DeepNetQoE framework capable of effectively guiding the training process with limited computing resources.
  \item \textbf{Builds the user's experience model}. By analyzing factors influencing the model training and quality of experience, a user's experience model is built to evaluate the experience value of different models to help the user choose model training.
  \item \textbf{Constructs the optimization of resources under multiple deep learning tasks}. The article proposes a resource optimization plan for multi-model training applications based on the QoE model and presents solutions.
  \item \textbf{Verifies the effectiveness of DeepNetQoE based on the crowd counting model}. The article verifies the feasibility and effectiveness of the proposed plan based on the crowd counting model.
\end{itemize}

The remainder of this article is organized as follows. Section II builds the DeepNetQoE architecture and illustrates it in a typical application. Section III discusses metrics to evaluate the performance of the DeepNetQoE. Section IV builds a QoE model based on the DeepNetQoE and conducts evaluation experiments and Section V concludes the article.

\section{DeepNetQoE Architecture and Typical Application Scenario}
This section builds the DeepNetQoE framework for training in a deep network model to guide users to make effective decisions. It is based on investing reasonable computing resources during training to gain a better training experience. In addition, the article introduces a typical scenario to which DeepNetQoE can be applied.

\subsection{DeepNetQoE Architecture}
Figure 2 shows a DeepNetQoE framework oriented on a deep network model. The framework consists of four layers: training layer, prediction layer, QoE model layer, and estimation layer.

\subsubsection{Training layer}
The transfer and interaction of information among different layers of DeepNetQoE is realized on the premise that the deep learning model has gone through a period of training. At the training layer, data and codes are uploaded to the server through a communication network and are trained on the GPU. With the increase of iterations, optimal models are generated constantly, and loss and evaluation values are generated in each epoch round at the same time. The change in loss indicates the degree of the model¡¯s convergence, while the evaluation value presents the model's performance. After a certain epoch is reached, this data is sent to the prediction layer.
\subsubsection{Prediction layer}
After receiving the losses and evaluation values in a time series, the prediction layer will predict loss and the evaluation value for future epochs. This is done through the use of the deep network's sequential data modeling. The predicted loss value and evaluation value will reflect the convergence and performance of the later training model. We predict the model performance under a time series by using the Long Short-Term Memory (LSTM) network due to its good sequence modeling performance and use the predicted losses to guarantee model convergence. The evaluation result is transmitted to the estimation layer as the input to the self-adaptive QoE model.
\subsubsection{QoE Model layer}
DeepNetQoE focuses on the user's different requirements in the process of training and use of the deep learning model. Thus, the QoE model layer aims to build a self-adaptive QoE model based on different model and user conditions. Variants used to build the self-adaptive QoE model are associated with multiple factors and depend mainly on users¡¯ needs to realize self-adaption. The most important factor for all users is performance, which directly decides the QoE of the model. Another is that optimization of computing resources is critical but can be different dependent on availability. Other factors include space complexity and testing time, etc. The self-adaptive QoE model will generate metrics to measure the experience value toward a deep learning model under the influence of multiple factors. It should be noted that the experience value is the result of the QoE model. In return, the user may set an expected weight parameter to solve the optimal estimation of computing resource consumption and space complexity among other metrics.
\subsubsection{Estimation layer}
The results generated by the estimation layer have an important link to the DeepNetQoE, which are used  to guide the model training. Optimal times of epoch and experience values under certain restrictions are  obtained from the data delivered by the prediction layer and the model built by the QoE model layer. The training iteration time when the experience value reaches the higher, along with the increase of iterations and the changes in QoE in the future, are  obtained. This is based on the computing resources and the expected experience value of the user. Finally, the result will be reported to the training layer to effectively regulate the training of the deep model.

DeepNetQoE is a loop-locked dynamic interaction system. In the process of deep model training, the four layers interact with each other constantly and adjust training strategies ceaselessly. The strategy to obtain the maximum experience value is adopted on the basis of satisfying the personalized needs of each user, and finally the adaptive optimal network model is acquired.
\subsection{Application of DeepNetQoE on Crowd Counting}
DeepNetQoE can be used in most of the deep network models to guide model training. The article will take the crowd counting model as an example to study the self-adaptive QoE model in deep networks. Crowd counting has a wide range of applications, such as estimating the number of participants in social and sports events. The common method for crowd counting is a deep network that processes the image to a density map.  The crowd counting will be then estimated by a summation over the predicted density map. In addition, there are some typical models with better performance on a large-scale crowd dataset. For instance, a Multi-Column Convolutional neural network (MCNN) is used to extract head features of different scales~\cite{MCNN}. Other models such as CSRNet~\cite{CSRNet} and SANet~\cite{SANet} have similar network structures to acquire crowd counting. It is worth noting that Ma et al.~\cite{BL} propose Bayesian Loss, a novel loss function, which constructs a density contribution probability model from the point annotations. In addition, there are also some pre-trained models, such as VGG, Alexnet and Res50~\cite{C3F} which can also be used for computational tasks. We will select some of the representative models to evaluate the performance of the curve prediction and the verification of resource allocation algorithms over time.

\section{Performance Metrics of DeepNetQoE}
In this section, key factors influencing QoE are described and the performance prediction plan for a specific model under time series is then illustrated simultaneously. DeepNetQoE will compute effective optimization and allocation of resources based on these factors and plans.
\subsection{Illustrations of DeepNetQoE Performance Metrics}
When training a deep network model on a GPU server, it is assumed that the trained model possesses full authority over the GPU's computing resources.  Moreover, the GPU will not load any other computational task during the training process. The article considers that the user's expectation on the model is influenced by multiple factors, as shown in Table I. A detailed introduction will be described next.


\begin{table*}
\caption{Definition of DeepNetQoE Performance Metrics}
\begin{center}
\begin{tabular}{l|l} \hline
\textbf{Performance Metrics}          & \textbf{Definition} \\ \hline
\ $e_{mae}$ & An important network performance metrics influencing the user's experience value in regression task\\
\ &associated with MAE, which is defined by natural index $exp$ and normalized parameters. \\
\ $e_{mse}$ & An important network performance metrics influencing the user's experience value in regression task \\
\ &associated with MSE, which is defined by natural index $exp$ and normalized parameters. \\
\ $e_{train}$ & A critical metrics to be considered when computing resources are limited, which occupies most of \\
\ &computing resources and influences the experience value of model training directly, associated with the network \\
\ &model training time $t_{train}$. \\
\ $e_{load}$ & A metrics influencing the user's experience value, a tiny influencing factor, which occupies instant\\
\ & computing resources, associated with the loading time $t_{load}$ of the network model. \\
\ $e_{test}$ & A metrics influencing user's experience value when deploying and using the network model, which occupies\\
\ & real-time computing resources, and shows different influence for different tasks, associated with the test time \\
\ & $t_{test}$ of the network model. \\
\hline
\end {tabular}
\label{Definition}
\end{center}
\end{table*}

The main objective of the iterative training of a deep network on a dataset is to optimize the weighting parameters of the neuron to make it constantly fit the features of the dataset and entail brings a high precision rate on the testing dataset. Therefore, a top factor influencing the experience value of the network model is the precision rate of the model. Different network models and tasks adopt different evaluation methods. Evaluation methods used for regression tasks and crowd counting include mean absolute error (MAE) and the mean squared error (MSE). The QoE model associated with the performance of these two indicators for measuring the performance of the mode is expressed as $e_{mae}$ and $e_{mse}$ respectively.

In addition to the performance of the model, another factor influencing experience value is the consumption of resources. Full occupation over the GPU is assumed in a single-model training process. Thus, this article takes the training time to indicate the consumption of resources. Here, there are three types of time-scales considered including the training time, loading time, and testing time of the model, expressed as $t_{train}$, $t_{load}$ and $t_{test}$, and the QoE model associated with them are expressed as $e_{train}$, $e_{load}$ and $e_{test}$, respectively. Models are constructed based on the user's expectation of these times. According to QoE analysis, among the acceptability of $t_{train}$, $t_{load}$ and $t_{test}$, the latter directly influences the real-time response performance of the model through an end-to-end test, which is critically important as an oversized $t_{test}$ leads to an undersized QoE. $t_{load}$ is used for the time spent on loading the model and thus, has loose requirements. The $t_{train}$,  required for model training is great in most cases. However, if the desired optimal model requires a huge consumption of computing resources, the user's tolerance will approach a limit and the experience value will decrease. Models for $e_{train}$, $e_{load}$ and $e_{test}$ are built based on the influence of different factors on QoE.
\subsection{Early Prediction of DeepNetQoE Performance via LSTM}
To obtain the total experience value $E_{all}$, individual¡¯s experience values of $e_{mae}$, $e_{mse}$, $e_{train}$, $e_{load}$ and $e_{test}$ need to be clearly defined. Factors influencing experience values are related performance and time, which can only be obtained during the training process. Since the training process of the model is changeable dynamically, performance and convergence degree of models in the future are full of uncertainties. To obtain experience values, it is necessary to predict curve dynamics based on the data generated in the training process. Therefore, it is important to accurately predict the curve of performance changing with the time. During the model's training process, a new round of training is conducted with an epoch as the node. Therefore, the number of epochs is used to replace the training time $t_{train}$. To a certain degree, the number of epochs can be mapped as $t_{train}$. Moreover, for a specific model, the values of $t_{test}$ and $t_{load}$ are confirmed and will be imported directly without prediction.

\begin{figure}
\centering
\subfigure[Loss prediction and Error]{
\begin{minipage}[b]{0.48\linewidth}
\includegraphics[width=1\linewidth]{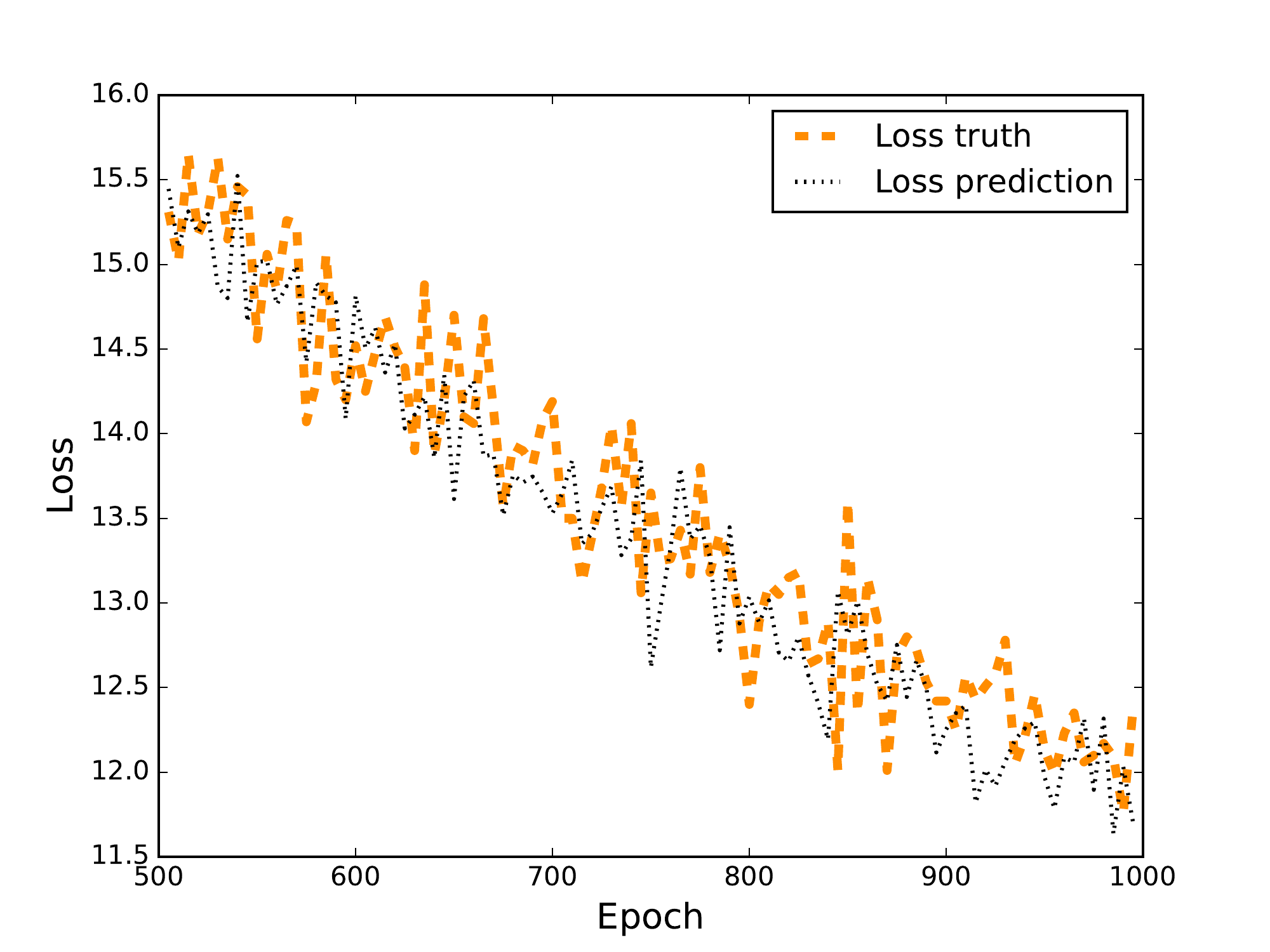}\vspace{4pt}
\includegraphics[width=1\linewidth]{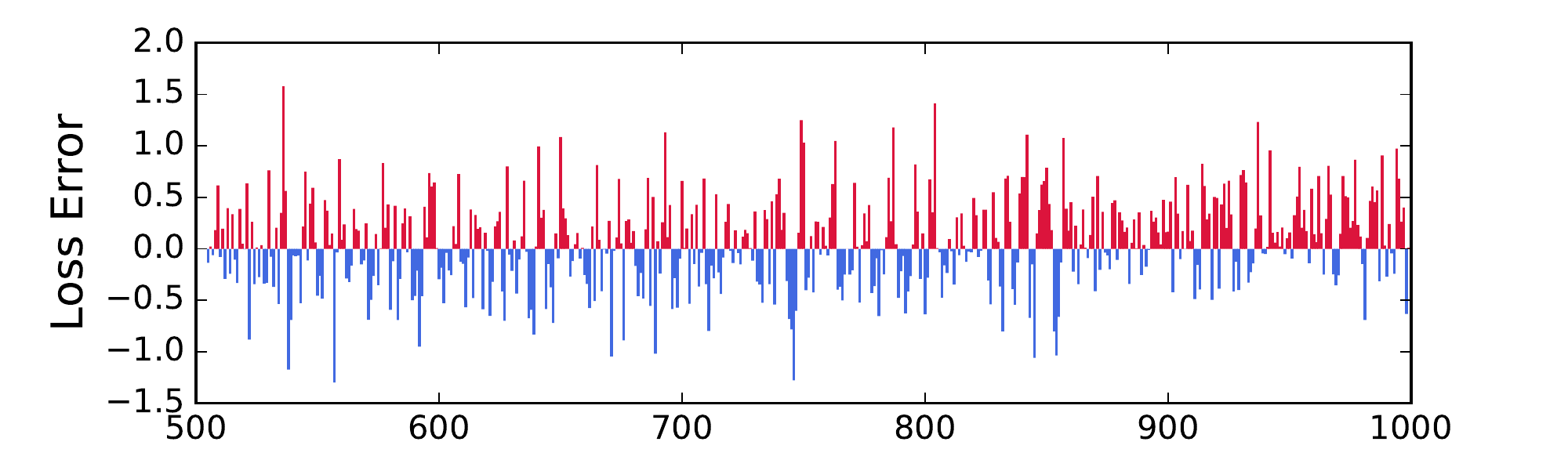}\vspace{4pt}
\end{minipage}}
\subfigure[MAE prediction and Error]{
\begin{minipage}[b]{0.48\linewidth}
\includegraphics[width=1\linewidth]{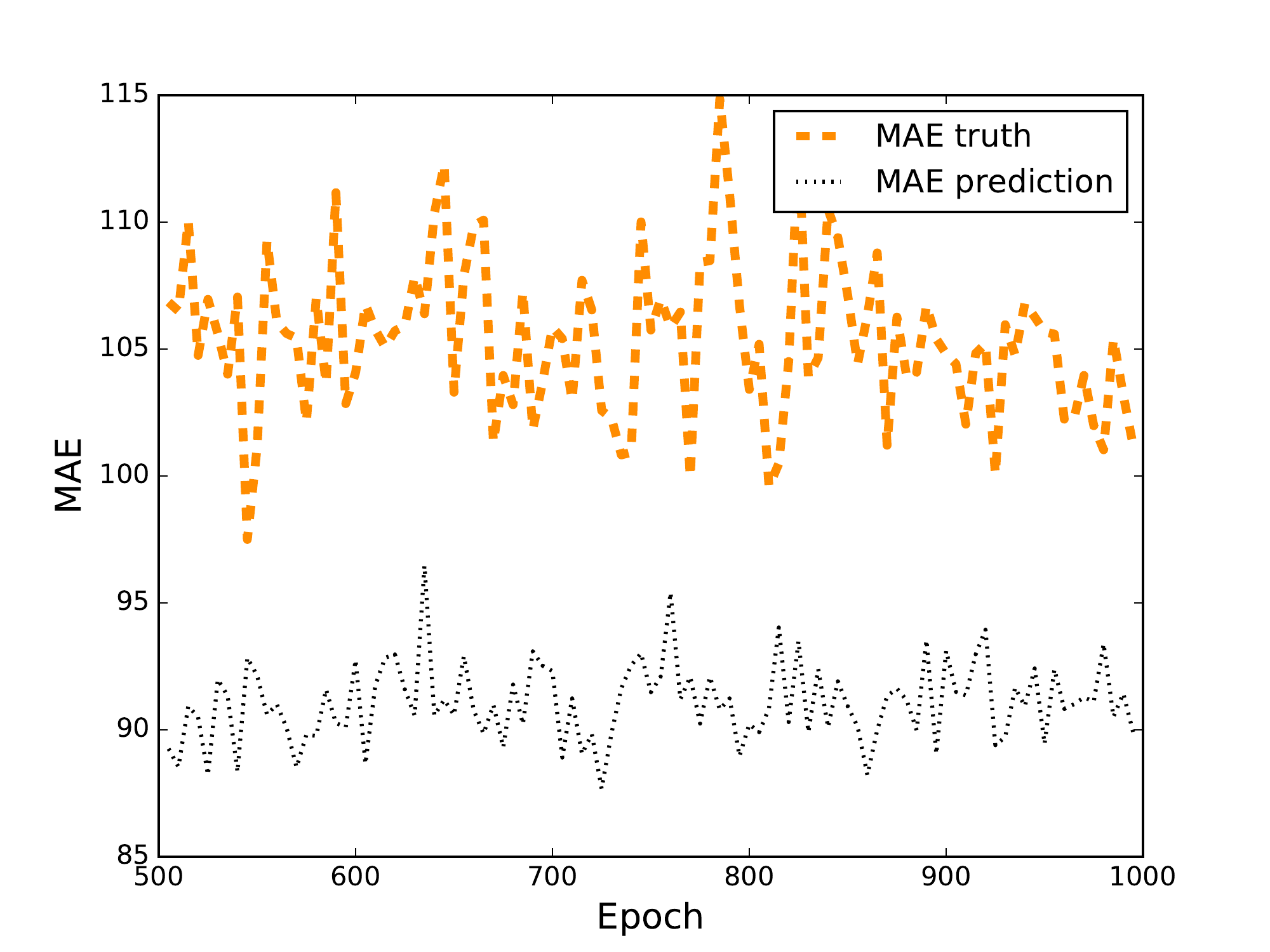}\vspace{4pt}
\includegraphics[width=1\linewidth]{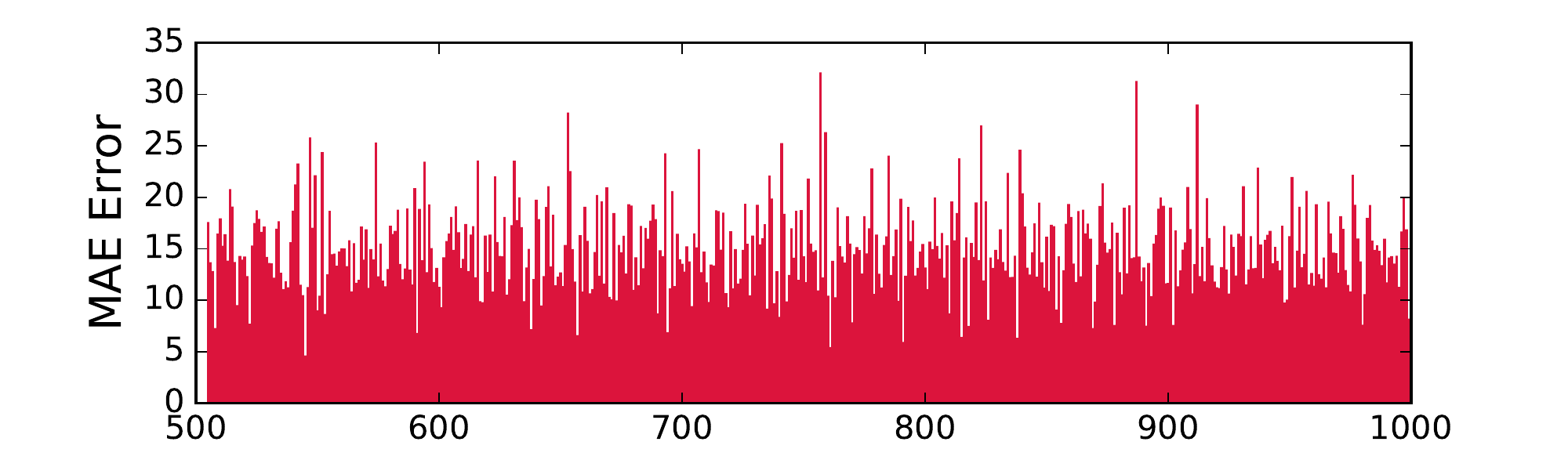}\vspace{4pt}
\end{minipage}}
\caption{Some experiment results from LSTM network to predict performance in time series.}
\label{prediction}
\end{figure}

The network structure used in this article is a two-layer LSTM network, the ReLU activation function is used for de-linearization and dropout is used for de-overfitting~\cite{lstmXu}. Generally, the epoch of the crowd counting model is 1,000, so it will produce 1,000 performance index results and loss values. We use the data from the first 500 epochs as the training dataset, and the data from the last 500 epochs as the test dataset. Figure 3 shows the experimental results of the LSTM model on the Bayesian Loss model. Figure 3 (a) displays predicted and real losses while Figure 3 (b) shows MAE and the difference between the predicted value and the real value. The experimental results prove that LSTM model has high prediction accuracy performance and the predicted results can be used for experience value analysis.

\section{DeepNetQoE Model and Performance Evaluation}
Combining the previously mentioned factors influencing the QoE model and the performance prediction under a time series model, a QoE model is established.  Then, the problems concerning the allocation of resources under multiple deep model training tasks are proposed. Based on the deep learning model for crowd counting, related experiments are subsequently conducted to verify the effectiveness and necessity of the proposed scheme.

\subsection{DeepNetQoE Model and Self-adaptive QoE Optimization}
Based on the performance evaluation metrics of DeepNetQoE, a complete QoE model based on the deep learning model can be obtained as:
\begin{equation}\label{eall}
\begin{aligned}
  &E_{all}(e_{mae},e_{mse},e_{train},e_{load},e_{test}) = \sum_{i=1,j}^{M=5,E}\omega_i\cdot e_j
\end{aligned}
\end{equation}
where $E = \{e_{mae}, e_{mse}, e_{train}, e_{load}, e_{test}\}$, $e_j\in E$, $M$ represents these five factors, $\omega_i$ means the user's expected weight on the $i_{th}$ factor and the expected weight will be determined in view of the personal experience and conditions of the user. When modeling each factor, the experience value of each factor will be limited $(0,1]$. $E_{all} (e_{mae}, e_{mse}, e_{train}, e_{load}, e_{test})\in (0,1]$ can be obtained based on the features of $e_{mae}$, $e_{mse}$, $e_{train}$, $e_{load}$ and $e_{test}$. Therefore, with the QoE of a single network model, each user by setting $E_{all}$ and and expected weight, can get the corresponding $e_{mae}$, $e_{mse}$, $e_{train}$, $e_{load}$, and $e_{test}$ in order to produce individualized requirements.

Allocation of resources for multi-network models aims to improve the total experience of a user engaged in joint training of multiple models. The user has different experience values for different models at the time of model series training, thus optimization problems of the total experience value ($E_{all}^{sum}$) can be obtained. The purpose is to maximize $E_{all}^{sum}$ via the allocation of computing resources according to the following definition:
\begin{equation}\label{esum2}
\begin{aligned}
& \mathop{\arg\max}_{e_{pref}^u,e_{train}^u,e_{load}^u,e_{test}^u} \sum_{u}^{\mathcal R}E_{all}^u \\
& \begin{array}{r@{\quad}r@{}l@{\quad}l}
s.t.& e_{mae}^u,e_{mse}^u  = f(t_{train}^u),& t_{train}^u\in e_{train}^u\\
& \sum_u^{\mathcal{R}} t_{train}^u  \leq \mathcal{T},& t_{train}^u\in e_{train}^u  \\
\end{array}
\end{aligned}
\end{equation}
Where $E_{all}^u$ refers to the experience value of the $u_{th}$ network model. Similarly, the influencing factor of the $u_{th}$ model is $e_{mae}^u$, $e_{mse}^u$, $e_{train}^u$, $e_{load}^u$, and $e_{test}^u$. $\mathcal R$ refers to the network model set to be trained, whereas $e_{mae}^u$ and $e_{mse}^u$ represents the performance of the $u_{th}$ model. $t_{train}^u$ corresponds to the training time of the $u_{th}$ model. Bear in mind that $e_{mae}^u$ and $e_{mse}^u$ are determined by the epoch of training, while the epoch can be mapped as a function of time. $f (\cdot)$ represents the mapping relation from $t_{train}^u$ to $e_{mae}^u$ and $e_{mse}^u$. As described in Section III, we can then obtain the predicted epoch-performance curve. $\mathcal{T}$ refers to the total training time possessed by a user and the training time is regarded as a computing resource. The value of $t_{load}^u$ and $t_{test}^u$ of each model is fixed, hence can be directly imported. Furthermore, loading and testing time occupy little computing resources when compared to the training time and therefore are excluded from the total training time $\mathcal{T}$.

According to the resource allocation and optimization model established above, the problem needs to be solved to maximize the user experience value under limited computing resources. First, an LSTM-based model is used to predict performance. This is based on the historical calculation results of the model to obtain the performance value of the subsequent epoch. The experience value of each model, under the corresponding epoch and required training time, are calculated. We solve the problem of computing resource allocation for the model based on a genetic algorithm. The genetic algorithm is utilized to generate the epoch of each model that needs to be trained. This includes conversion into a binary string, and then through cycles of individual elimination, selection, hybridization, and mutation several times to obtain all the results satisfying restrictions. Finally, under the limitation of the total training time $\mathcal T$, the total experience value, $E_{all}^{sum}$, is obtained and the training epoch of each network model is used to guide the user's resource allocation.

\subsection{DeepNetQoE Performance Evaluation}
By combining the user's QoE model and the computing resources allocation algorithm, we carry out experiments. We analyze the results to verify the effectiveness of the model. In setting up these experiments, only one network model is used during training and there are no other unnecessary processes on the server. Under these conditions, the utilization rate of the network model to the server is above 85\%.


\begin{figure*}
\centering
\subfigure[Total experience value from different w.]{
\begin{minipage}[b]{0.31\linewidth}
\includegraphics[width=1\linewidth]{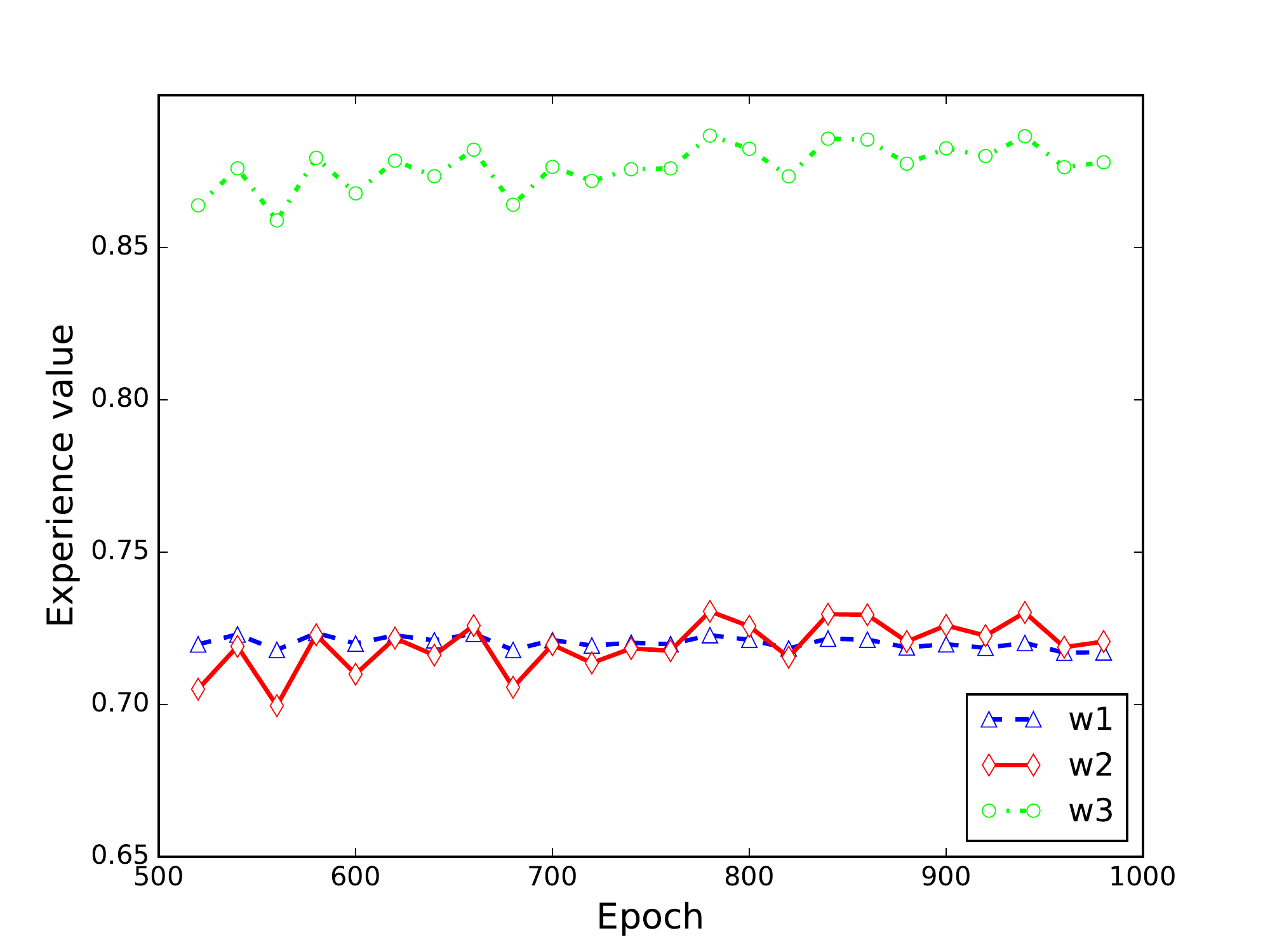}\vspace{4pt}
\end{minipage}}
\subfigure[Total experience value from different algorithms.]{
\begin{minipage}[b]{0.31\linewidth}
\includegraphics[width=1\linewidth]{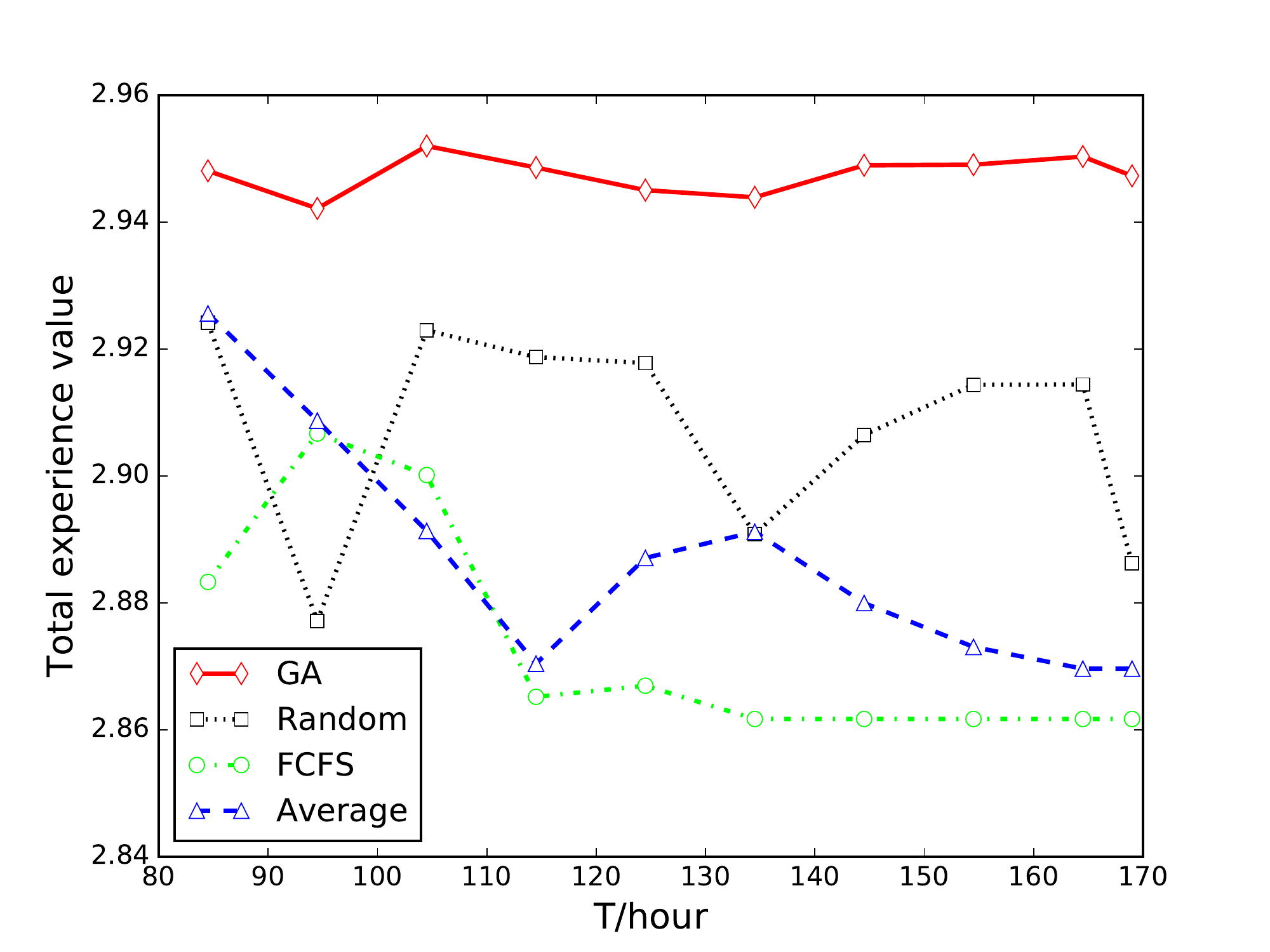}\vspace{4pt}
\end{minipage}}
\subfigure[Different model experience value from GA.]{
\begin{minipage}[b]{0.31\linewidth}
\includegraphics[width=1\linewidth]{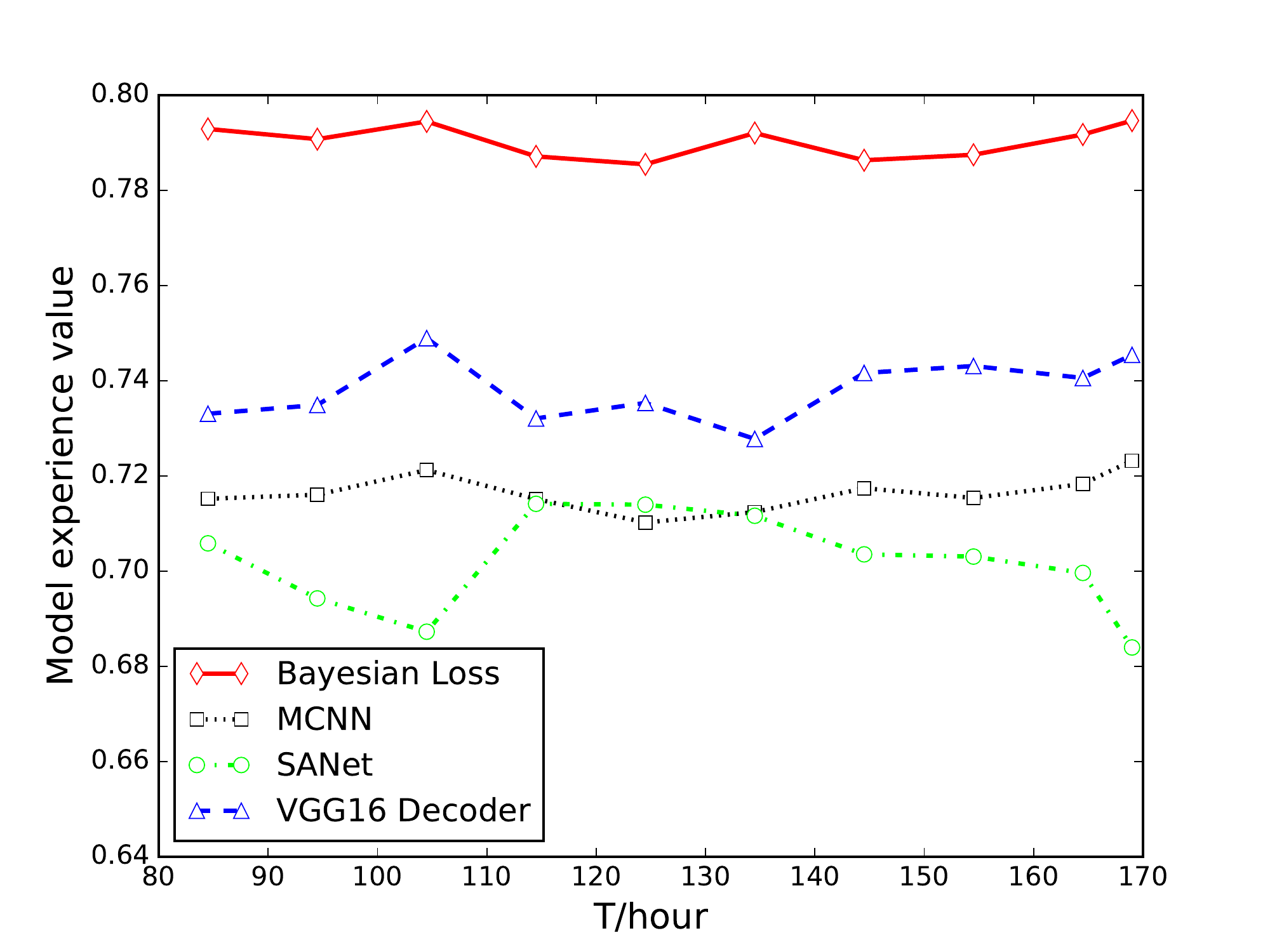}\vspace{4pt}
\end{minipage}}
\caption{Results of self-adaptive experience value and optimization.}
\label{optimizationmulti}
\end{figure*}

Four networks, including Bayesian loss, MCNN, SANet, and VGG16 Decoder, show good performance in the crowd counting task. These models have been trained through 1,000 iterations on the GPU of Navida V100\_Group. The video memory size is 32 Gigabyte(GB) while the internal memory is 128 GB.  The data set used for the model training is UCF-QNRF with massive crowd. Parameters associated with the network model are shown in Table II. In this table MAE and MSE are evaluation metrics of the optimal model to help each user to decide the final experience value.

\begin{table}[h]
\caption{The parameters of crowd counting model}
\begin{center}
\begin{tabular}{l|l|l|l|l|l}
\hline
Method& MAE & MSE & TrainT &TestT & LoadT\\ \hline \hline
BL\cite{BL}&89.38&161.67&14 h&0.2335 s&15 s       \\
MCNN\cite{MCNN}&185.86&287.15&30 h&0.3353 s&17 s        \\
SANet\cite{SANet} &129.91&217.39&50 h&0.8294 s&16 s \\
VGG16 Decoder\cite{C3F} &145.97&247.94&35 h&0.6647 s&18 s\\\hline
\end{tabular}
\end{center}
\label{tableII}
\end{table}

Furthermore, when assessing the user's experience value, it is necessary to determine parameters of the model according to the training parameters of the network itself to realize normalized processing on the QoE models. In the experiment of crowd counting described earlier, QoE model parameters are obtained from the constant debugging of the actual parameters of the network model. It should be noted that parameter, $omega$, is the expected weight of the user, which is determined by the user's personal needs.

Based on the experience value model, two aspects of experiments were carried out. Figure 4(a) shows the impact of different expected weights, w, on the QoE model based on the Bayesian loss model. Among them, $w_1 = [0.1, 0.1, 0.5, 0.05, 0.25]$, $w_2 = [0.4, 0.4, 0.05, 0.03, 0.12]$, $w_3 = [0.3, 0.4, 0.01, 0.2, 0.09]$. On the other hand, with the goal of optimizing Equation (\ref{esum2}), the total experience value when multiple models coexist is solved. Figure 4 uses four algorithms to perform resource allocation tasks. The allocation of computing resources is based on 500 basic iterations of each model. Note that the GA method is based on the genetic algorithm in \cite{KoopialipoorGA}. Random method allocates the remaining resources to each network model after allocating the basic resources, and then randomly allocates them. The FCFS method uses the concept of first-come-first-served. After the basic resources have been allocated, the remaining resources are used allocated in the order of Bayesian loss, MCNN, SANet, and VGG16 Decoder. Each model stops training after reaching the maximum number of iterations of 1000 times. Average method is to distribute resources equally to each network model. From Figure 4(b), we can see that the GA method shows the best performance, and the total experience value of the allocation scheme is maintained above 2.94 under different total computing resource settings. The Random method shows a good performance on some computing resources. The performance of the FCFS method is relatively poor. Figure 4 (c) displays the experience value generated by the GA method for the four models on different total computing resources. It indicates that the Bayesian loss and VGG16 decoder have the highest experience value.
\section{Conclusion}
The deep network model training process consumes a lot of computing resources and has unknown situations, this article proposes a self-adaptive QoE optimization framework of deep networks.  It also builds a computing resource optimization scheme based on user requirements and total computing resources. A DeepNetQoE framework for deep network model training is constructed to guide users to make effective decisions during the training process. Later, five evaluation factors of DeepNetQoE which influence the QoE model are described. Based on a genetic algorithm and targeted at maximizing the total experience value, we allocated limited resources to the training of each network model. By taking the four models in the crowd counting network as an example, we conducted several experiments. The results prove the advancement and strong adaptivity of DeepNetQoE in the training of deep learning networks.


\section*{Acknowledgment}
This work is supported by the National Key R\&D Program of China (2017YFE0123600), and Shenzhen Institute of Artificial Intelligence and Robotics for Society (AIRS).

\bibliographystyle{IEEEtran}
%

\end{document}